\documentclass[conference]{IEEEtran}
\IEEEoverridecommandlockouts
\usepackage{cite}
\usepackage{amsmath,amssymb,amsfonts}
\usepackage{algorithmic}
\usepackage{graphicx}
\usepackage{textcomp}
\usepackage{xcolor}
\usepackage{url}
\usepackage{multicol}
\usepackage{multirow}
\usepackage{float}
\usepackage{array}
\usepackage{hyperref}
\hypersetup{
    colorlinks=true,
    linkcolor=blue,     
    citecolor=blue,     
    urlcolor=blue,    
    anchorcolor=blue,   
}

\def\BibTeX{{\rm B\kern-.05em{\sc i\kern-.025em b}\kern-.08em
    T\kern-.1667em\lower.7ex\hbox{E}\kern-.125emX}}
\begin{document}

\title{UltraGS: Real-Time Physically-Decoupled Gaussian Splatting for Ultrasound Novel View Synthesis\\
\thanks{This work was supported by the National Natural Science Foundation of China (Grant No. 62306003), the Open Research Fund of Guangdong Laboratory of Artificial Intelligence and Digital Economy (SZ) (Grant No. GML-KF-24-29), and the Open Foundation of Jiangxi Provincial Key Laboratory of Image Processing and Pattern Recognition (Grant No. ET202404437).}
}
\author{
\IEEEauthorblockN{
Yuezhe Yang\textsuperscript{1},
Qingqing Ruan\textsuperscript{2},
Wenjie Cai\textsuperscript{1},
Yufang Dong\textsuperscript{3},
Dexin Yang\textsuperscript{1},
Xingbo Dong\textsuperscript{1}\textsuperscript{\textdagger},
Zhe Jin\textsuperscript{1},
Yong Dai\textsuperscript{4}
}

\IEEEauthorblockA{\textsuperscript{1}
Anhui Provincial International Joint Research Center for Advanced Technology in Medical Imaging, Anhui University, Hefei, China\\
\{wa2214014, wa2214030, y82214109\}@stu.ahu.edu.cn, \{xingbo.dong, jinzhe\}@ahu.edu.cn}

\IEEEauthorblockA{\textsuperscript{2}
Xinhua Hospital Affiliated to Shanghai Jiao Tong University School of Medicine, Shanghai, China\\
ChristinaRuan111@outlook.com}

\IEEEauthorblockA{\textsuperscript{3}
School of Medicine, Nankai University, Tianjin, China\\
dongyufang@mail.nankai.edu.cn}

\IEEEauthorblockA{\textsuperscript{4}
School of Medicine, Anhui University of Science \& Technology, Huainan, China\\
daiyong22@aust.edu.cn}

\thanks{\textdagger\ Corresponding author: Xingbo Dong (xingbo.dong@ahu.edu.cn).}
}

\maketitle

\begin{abstract}
Ultrasound imaging is a cornerstone of non-invasive clinical diagnostics, yet its limited field of view poses challenges for novel view synthesis. We present UltraGS, a real-time framework that adapts Gaussian Splatting to sensorless ultrasound imaging by integrating explicit radiance fields with lightweight, physics-inspired acoustic modeling. UltraGS employs depth-aware Gaussian primitives with learnable fields of view to improve geometric consistency under unconstrained probe motion, and introduces PD Rendering, a differentiable acoustic operator that combines low-order spherical harmonics with first-order wave effects for efficient intensity synthesis. We further present a clinical ultrasound dataset acquired under real-world scanning protocols. Extensive evaluations across three datasets demonstrate that UltraGS establishes a new performance-efficiency frontier, achieving state-of-the-art results in PSNR (up to 29.55) and SSIM (up to 0.89) while achieving real-time synthesis at 64.69 fps on a single GPU. The code and dataset are open-sourced at: \url{https://github.com/Bean-Young/UltraGS}.
\end{abstract}


\section{Introduction} 
\label{sec:intro} 
Ultrasound imaging is highly valued in clinical practice for its non-invasive nature, real-time capabilities, and cost-effectiveness \cite{yang2025annotated}. Despite these advantages, ultrasound suffers from a limited scanning angle, preventing the acquisition of a complete field of view (FoV) in a single frame \cite{adriaans2024trackerless}. This limitation complicates the construction of reliable 3D models essential for diagnosis and treatment planning. Furthermore, ultrasound data exhibits anisotropic characteristics, leading to view-dependent inconsistencies that adversely affect novel view synthesis (NVS) \cite{dagli2024nerf}. Traditional solutions, such as tracked probes or mechanical sweepers, rely heavily on manual alignment or expensive hardware, making them impractical for widespread clinical adoption \cite{mozaffari2017freehand}.

Computer-assisted approaches have attempted to bridge this gap. Early sensorless methods aimed to reduce costs but often failed to accurately model anatomical structures \cite{housden2007sensorless}. More recently, deep learning techniques have leveraged data-driven priors to improve synthesis quality \cite{luo2022deep,chen2023freehand}. However, these methods are often limited in modeling complex geometries from arbitrary viewpoints and face challenges related to privacy protection and domain shifts \cite{guan2021domain}.

The advent of NeRF \cite{mildenhall2021nerf} introduced a powerful implicit representation for NVS. Ultra-NeRF \cite{wysocki2024ultra} further adapted this framework by integrating ultrasound physics to produce physically consistent B-mode images. However, NeRF-based methods rely on computationally expensive volume ray marching, which prevents real-time performance—a critical requirement for medical imaging. To address this efficiency bottleneck, 3D Gaussian Splatting (3DGS) \cite{kerbl20233d} was introduced, enabling real-time rendering via explicit Gaussian primitives. While 3DGS has been successfully adapted for X-ray CT reconstruction \cite{cai2024radiative}, its direct application to ultrasound is non-trivial due to fundamental physical differences. Unlike transmission-based X-rays, ultrasound relies on complex wave propagation phenomena, including scattering, reflection, and attenuation. Consequently, standard Gaussian accumulation cannot accurately describe this physical process.

Applying 3DGS to ultrasound imaging encounters three fundamental challenges. First, the lack of explicit depth cues. While implicit depth encoded in Gaussian positions suffices for optical scenes , ultrasound demands the explicit modeling of tissue-probe interfaces and acoustic interactions to recover metric-consistent structural depth. Second, the dynamic imaging aperture. Unlike the fixed intrinsic parameters of optical lenses, the effective FoV in ultrasound fluctuates dynamically due to depth-dependent imaging ranges, adaptive beam focusing, and unconstrained probe motion. Third, ultrasound intensity exhibits strong depth-dependent attenuation and speckle patterns, which are costly to simulate with full physical models and difficult to capture using purely data-driven approaches.

To address these challenges, we propose UltraGS, a real-time framework tailored for sensorless B-mode ultrasound under unconstrained clinical scanning. By modeling ultrasound acquisition as a sequence of planar slices and adopting a virtual pinhole proxy for pose optimization, UltraGS leverages Gaussian Splatting to recover a globally consistent 3D representation from freehand scans. The framework incorporates depth-aware Gaussian primitives with learnable fields of view to improve geometric stability, together with a lightweight, differentiable approximation of acoustic wave effects for efficient intensity synthesis. Our key contributions are as follows: 

1) We present UltraGS, a practical adaptation of 3D Gaussian Splatting for sensorless ultrasound novel view synthesis, enabling real-time reconstruction under unconstrained clinical scanning conditions.

2) Technically, we formulate a depth-aware Gaussian splatting strategy with Dynamic Aperture Rectification (DAR) via learnable fields of view. We develop Physically-Decoupled (PD) Rendering, a differentiable neural acoustic operator that couples low-order spherical harmonics with first-order wave physics including attenuation, reflection, and scattering. 

3) We introduce the Clinical Ultrasound Examination Dataset, a novel benchmark consisting of diverse anatomical scans acquired under rigorous real-world clinical protocols. 

4) Extensive evaluations demonstrate that UltraGS achieves state-of-the-art (SOTA) reconstruction quality (PSNR up to 29.55) while delivering unprecedented real-time synthesis at 64.69 fps.

\section{Related Work}

The landscape of NVS has been fundamentally reshaped by NeRF \cite{mildenhall2021nerf}, which utilize implicit MLPs to represent continuous scene geometry. While advancements in anti-aliasing  \cite{barron2021mip} and baking strategies \cite{chen2023mobilenerf}  have improved inference speed, the hereditary latencies of volume ray marching remain a significant barrier to real-time clinical deployment \cite{wysocki2024ultra}. To circumvent these bottlenecks, 3DGS \cite{kerbl20233d} introduced an explicit radiance field using anisotropic primitives, enabling high-speed rasterization. Subsequent extensions have refined this framework through surface-alignment \cite{guedon2024sugar}, frequency-aware filtering \cite{yu2024mip}, and depth-aware priors \cite{kumar2024few}. However, these paradigms are primarily optimized for optical light transport and fail to account for the complex acoustic interactions inherent in ultrasound imaging.

In the medical domain, NVS is essential for volumetric reconstruction and diagnostic navigation \cite{yang2025explicit}. Early efforts, such as Ultra-NeRF \cite{wysocki2024ultra}, successfully integrated ultrasound-specific physics into implicit frameworks, yet struggled with the prohibitive computational overhead of neural integration. Following the success of 3DGS, radiative variants like X-Gaussian \cite{cai2024radiative} and DDGS-CT \cite{gao2024ddgs} adapted the rendering equation for transmission-based X-ray and CT modalities. Despite these strides, ultrasound presents a fundamentally different challenge due to its anisotropic wave propagation and stochastic speckle formation. Unlike transmission-based imaging, ultrasound intensity is governed by non-linear attenuation and boundary-specific reflections that standard Gaussian accumulation cannot represent.

Furthermore, sensorless 3D ultrasound reconstruction remains a long-standing challenge due to the absence of hardware tracking \cite{adriaans2024trackerless}. Traditional deep-learning approaches for pose estimation \cite{luo2022deep, chen2023freehand} often rely on rigid data-driven priors, which struggle to generalize across diverse clinical protocols \cite{li2025tus}. Our work bridges this gap by introducing UltraGS, which reconciles explicit splatting with a PD Rendering paradigm. Unlike surface-aligned Gaussian methods designed for optical reflectance, UltraGS employs 2D Gaussian disks to explicitly model tissue–probe interaction planes. By formulating the synthesis process as a first-order approximation of acoustic wave physics and introducing Acoustic-Adapted Initialization, we achieve metric-consistent 3D reconstruction and high-fidelity intensity modeling without sacrificing real-time performance.

\section{Methods}
\begin{figure*}[t]
\centering
\includegraphics[width=\textwidth]{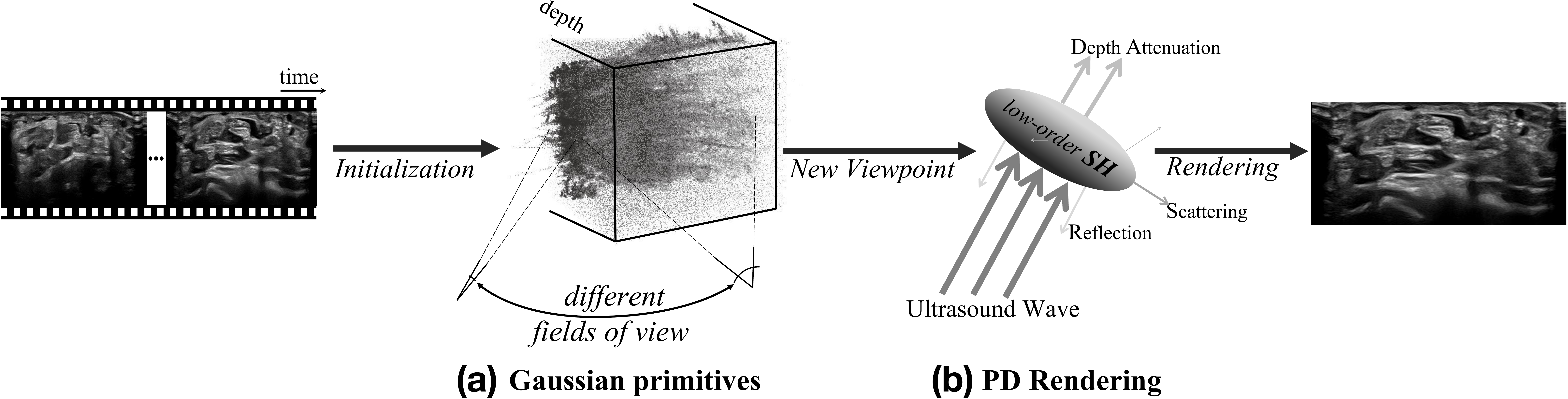}
\caption{Pipeline of UltraGS. (a) Gaussian primitives are assigned learnable fields of view for dynamic aperture rectification, enabling metric-consistent depth recovery. (Illustrated here is Case 1 from the Clinical Dataset.) (b) PD Rendering synthesizes new images by integrating low-order spherical harmonics with an efficient approximation of acoustic wave propagation (attenuation, reflection, and scattering).}
\label{fig-m2}
\end{figure*}

The UltraGS pipeline (Fig. \ref{fig-m2}) introduces two core components designed to address the geometric and photometric challenges of ultrasound novel view synthesis while maintaining real-time performance. \textbf{(1)} A depth-aware Gaussian primitive strategy that utilizes DAR to ensure metric-consistent structural reconstruction, and \textbf{(2)} PD Rendering, a differentiable neural acoustic operator that decouples view-dependent appearance from intrinsic wave-matter interactions. By coupling explicit geometry with first-order wave physics, this framework enables high-fidelity intensity modeling while circumventing the computational latencies inherent in traditional volumetric ray marching.

\subsection{Preliminaries of Gaussian Splatting}
In standard 3DGS, a scene is represented by a collection of anisotropic 3D Gaussians initialized from a sparse point cloud $\mathcal{P}$ obtained via Structure-from-Motion (SfM). Each primitive is defined by its center position $\mu$ and a covariance matrix $\Sigma$. During rendering, the world-space covariance is projected into camera-space $\Sigma^{\prime}$ using the Jacobian $J$ of the affine approximation of the projective transformation and the viewing transformation matrix $W$:
\begin{equation}
\Sigma^{\prime} = J W \Sigma W^\top J^\top. 
\end{equation}

\begin{figure}[t]
\centering
\includegraphics[width=0.45\textwidth]{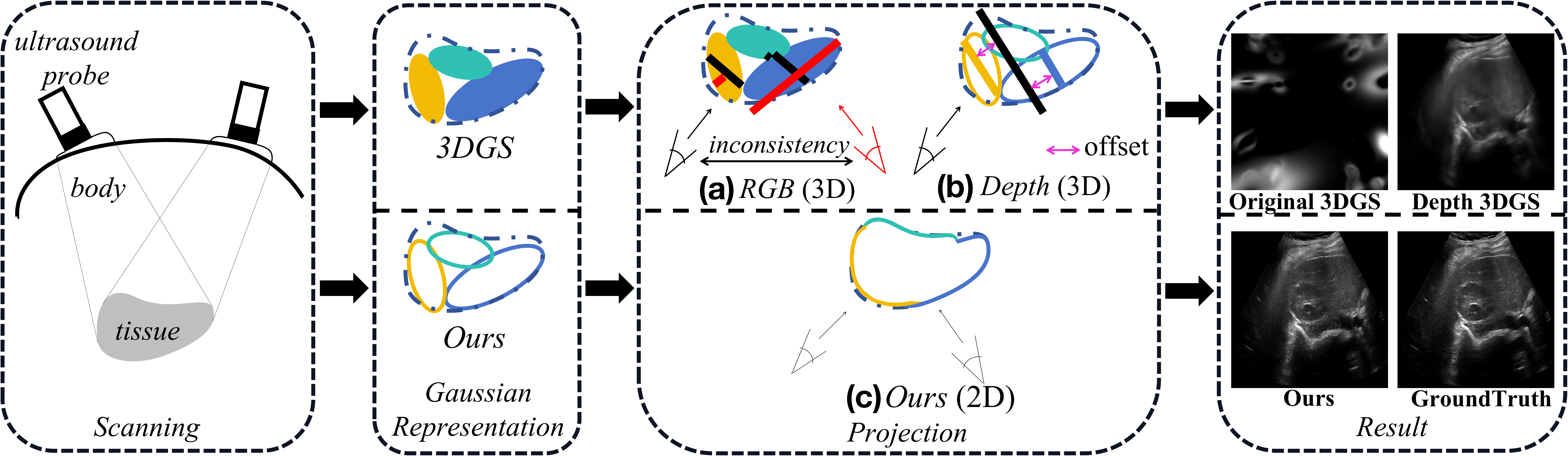}
\caption{Comparison of primitive representations. (a) Standard 3DGS suffers from view inconsistency and (b) significant depth offsets in acoustic scenes. (c) UltraGS utilizes 2D Gaussian disks to facilitate metric-consistent projection and superior structural fidelity.}
\label{fig-m1}
\vspace{-10pt}
\end{figure}

However, the reliance of 3DGS on optical viewpoint parameters introduces significant geometric instability in clinical ultrasound. Because traditional 3DGS formulations lack fixed normal directions, the projected Gaussian shapes vary inconsistently across different views, leading to rendering artifacts and substantial depth map offsets (Fig. \ref{fig-m1}(a, b)) \cite{huang20242d}. Such misalignments are detrimental to clinical diagnosis, which requires precise anatomical localization. While various depth-regularization techniques have been proposed, structural misalignment persists in acoustic scenes. To overcome these limitations, we utilize 2D Gaussian disks, which are well-suited to representing the tissue interfaces and boundaries captured by ultrasound. High-density stacking of these semi-transparent primitives yields a quasi-volumetric representation that more accurately models acoustic intensity variations in continuous media than isotropic 3D spheres, thereby ensuring superior geometric consistency and perspective-correct depth recovery (Fig. \ref{fig-m1}(c)).

\subsection{Metric-Consistent Depth Recovery and Dynamic Aperture Rectification}
To achieve hardware-free clinical accessibility, UltraGS operates under a strictly sensorless paradigm, eliminating the need for external tracking devices. We first implement an Acoustic-Adapted Initialization protocol. Recognizing that standard SfM is optimized for projective RGB geometry, we treat the output of COLMAP \cite{schoenberger2016sfm} not as absolute poses, but as stochastic global anchors. These anchors establish a baseline coordinate system while the subsequent optimization rectifies the optical-to-acoustic domain mismatch. 

Differing from volumetric 3D Gaussians, we utilize 2D Gaussian disks to model the prominent tissue-probe interfaces and structural surfaces within the imaging plane. We define a local 2D coordinate system using orthonormal tangent vectors $t_u$ and $t_v$. Each primitive is represented by its center $p_k$ and a $2\times2$ covariance matrix $\Sigma$. The 3D position $P(u,v)$ of any point on the primitive surface is formulated as:
\begin{equation}
P(u,v) = p_k + u \cdot t_u + v \cdot t_v,
\end{equation}
where the elliptical distribution in 3D space is governed by:
\begin{equation}
G(p) = \exp\left(-\tfrac{1}{2} [u, v] \Sigma^{-1} [u, v]^T \right).\end{equation}
To achieve high-fidelity depth recovery, we employ a Perspective-Correct Wave-Splat Intersection. For any image coordinate $(x, y)$, the pixel ray is parameterized as the intersection of two orthogonal planes $h_x = (-1, 0, 0, x)^T$ and $h_y = (0, -1, 0, y)^T$ in the camera coordinate system. These planes are transformed into the local primitive coordinates via $M = (WH)^{-1}$, where $H$ is the local-to-world transform and $W$ is the world-to-camera matrix. The intersection coordinates $(u, v)$ are evaluated to compute the homogeneous world coordinates $\mathbf{X}_w$:
\begin{equation}
\mathbf{X}_w = \begin{bmatrix} p_k + u \cdot t_u + v \cdot t_v \ 1 \end{bmatrix}.
\end{equation}
The final metric depth $z$ is extracted by transforming $\mathbf{X}_w$ into camera coordinates:
\begin{equation}
\mathbf{X}_c = W \cdot \mathbf{X}_w = \begin{bmatrix} X_c \ Y_c \ z \ 1 \end{bmatrix}.
\end{equation}
This explicit intersection ensures multi-view geometric consistency and perspective-correct depth recovery, addressing the artifacts observed in standard 3DGS. A pivotal component of our framework is  DAR via learnable fields of view. Unlike fixed-lens optical cameras, ultrasound effective FoV fluctuates due to beam focusing, depth-dependent imaging ranges, and unconstrained probe motion. To compensate for these geometric distortions and rectify initial SfM calibration errors, we introduce learnable parameters $\theta_x, \theta_y$. The rectified fields of view are defined as:
\begin{equation}
FOV_x = 2\tan^{-1}(e^{\theta_x}), \quad FOV_y = 2\tan^{-1}(e^{\theta_y}),
\end{equation}
which dynamically adjust the focal lengths $f_x, f_y$ in the intrinsic matrix $K$:
\begin{equation}
\begin{split}
K &= \begin{bmatrix} 
        f_x & 0 & \tfrac{W_{\text{img}}}{2} \\ 
        0 & f_y & \tfrac{H_{\text{img}}}{2} \\ 
        0 & 0 & 1 
    \end{bmatrix}, \\
f_{x,y} &= \frac{\text{dim}_{x,y}}{2\tan(\text{FOV}_{x,y}/2)}.
\end{split}
\end{equation}
By optimizing $\theta$ alongside Gaussian parameters through a neural-geometric feedback loop, UltraGS adaptively corrects initial pose drifts and intrinsic misalignments. This joint refinement ensures that the final 3D representation conforms to acoustic reality even when the initial anchors are suboptimal.

\subsection{PD Rendering: A Differentiable Neural Acoustic Operator}
In conventional 3DGS, spherical harmonic (SH) functions $Y_l^m(\theta, \phi)$ are employed to model view-dependent appearance, typically using a high order ($l=3$) to capture complex optical light transport. However, ultrasound imaging is dominated by wave-matter interactions rather than complex directional variations. Furthermore, the constrained viewing angles of clinical transducers reduce the necessity for high-order directional modeling. Consequently, we adopt a first-order SH representation ($l=1$) with 4 parameters to represent the base acoustic response $\mathbf{c}(d, k)$, providing a balance between representational capacity and real-time inference. To faithfully synthesize ultrasound intensities, we decouple the rendering process into three physically-grounded components: depth attenuation, specular reflection, and volumetric scattering.

1) Depth-Dependent Attenuation: 
Acoustic energy undergoes exponential decay as it propagates through tissue, governed by the Beer-Lambert law:
\begin{equation}
I(z) = I_0 e^{-\alpha z}
\end{equation}
where $I_0$ represents the initial acoustic intensity, $\alpha$ is the attenuation coefficient, and $z$ denotes the tissue penetration depth. However, clinical B-mode ultrasound systems typically apply logarithmic compression to the raw echo signals to fit their high dynamic range into a displayable intensity range.By transforming the physical decay into the log-compressed domain, we obtain $\ln(I(z)) = \ln(I_0 e^{-\alpha z}) = \ln(I_0) - \alpha z$.
Neglecting the constant initial offset $\ln(I_0)$, the intensity decay in the clinical display domain is linearly proportional to depth $z$. Therefore, we model attenuation as a differentiable linear proxy within the log-compressed domain:
\begin{equation}
\mathbf{I}_{att} = [-\alpha \cdot z, 0, 0]^\top
\end{equation}
where $\alpha$ is fixed to 1 in our implementation. In log-compressed B-mode ultrasound, the attenuation coefficient and display gain are not separately identifiable; fixing $\alpha$ therefore acts as a normalization choice without reducing the representational capacity of the model. This derivation ensures that our rendering pipeline aligns with the standard signal processing pipeline of clinical scanners.

2) Specular Reflection: Acoustic impedance mismatches at anatomical boundaries create specular reflections. We model this phenomenon through a quadratic intensity term that captures the interaction between the base reflectivity and the incident wave energy:
\begin{equation}
\mathbf{I}_{\text{refl}} = \beta  \mathbf{c} \odot \mathbf{c}, \text{}
\end{equation}
where $\beta$ is a learnable reflection coefficient and $\odot$ denotes the Hadamard product. This term ensures that tissue interfaces exhibit the characteristic brightness observed in clinical scans.

3) Volumetric Scattering: Sub-wavelength tissue structures produce diffuse returns known as scattering. We model these stochastic interactions using a learnable scattering matrix $\mathbf{\Gamma} \in \mathbb{R}^{3\times3}$ with zero diagonal entries to represent inter-channel coupling:
\begin{equation}
\mathbf{I}_{\text{scat}} = (\mathbf{\Gamma}  \mathbf{c}) \odot \mathbf{c}. \text{}
\end{equation}
4) Physically-Decoupled Composition: The final ultrasound intensity $\mathbf{I}_{\text{final}}$ is synthesized by a weighted combination of these components:
\begin{equation}
\mathbf{I}_{\text{final}} = \mathbf{c}(d,k) + w_{\text{att}} \mathbf{I}_{\text{att}} + w_{\text{refl}} \mathbf{I}_{\text{refl}} + w_{\text{scat}} \mathbf{I}_{\text{scat}}.
\end{equation}
The learnable weights $w_{att}$, $w_{refl}$, $w_{scat}$ are constrained to the positive domain via a Softplus function. This ensures adherence to physical principles of energy decay and reflection, preventing the optimization process from generating non-physical energy gain.

These weights allow the framework to adaptively balance physical effects across different anatomical regions while maintaining a constrained optimization space. By treating the rendering as a first-order physics approximation rather than a purely data-driven mapping, UltraGS achieves superior fidelity and real-time performance of 64.69 fps. Our goal is not to fully simulate acoustic wave propagation, but to identify a minimal set of first-order physical effects that are sufficient for stable geometry reconstruction and perceptually faithful B-mode synthesis under real-time constraints.

\begin{table}[tbp]
\centering
\scriptsize 
\caption{Quantitative comparison on Wild Dataset \& Phantom Dataset. }
\resizebox{0.5\textwidth}{!}{
    \begin{tabular}{
    >{\centering\arraybackslash}p{2cm}|
    *{3}{c}|
    *{3}{c}|
    >{\centering\arraybackslash}p{2cm}
    }
    \hline\hline
    \multirow{2}{*}{\textbf{Method}} & 
    \multicolumn{3}{c|}{\textbf{Wild Dataset}} &
    \multicolumn{3}{c|}{\textbf{Phantom Dataset}} &
    \multirow{2}{*}{\textbf{Speed (fps) $\uparrow$}} \\
    \cline{2-7}
    & \multicolumn{1}{c}{PSNR $\uparrow$} & \multicolumn{1}{c}{SSIM $\uparrow$} & \multicolumn{1}{c|}{MSE$\downarrow$}
    & \multicolumn{1}{c}{PSNR$\uparrow$} & \multicolumn{1}{c}{SSIM$\uparrow$} & \multicolumn{1}{c|}{MSE$\downarrow$}
     \\
    \hline
    NeRF & 20.176 & 0.6834 & 0.0066 & 20.359 & 0.6301 & 0.0058& 0.28   \\
    TensoRF & 24.058 & 0.7528 & 0.0051 & 27.260 & 0.7178 & 0.0029 & 1.61  \\
    Ultra-NeRF & 19.140 & 0.6218 & 0.0109 & 25.115 & 0.6814 & 0.0042 & 0.43 \\
    \hline
    3DGS & 22.327 & 0.7745 & 0.0057 & 27.115 & 0.7142 & 0.0031 & 52.56 \\
    SuGaR & 21.392 & 0.6291 & 0.0159 & 28.899 & 0.8843 & 0.0024 & 9.81 \\
    \textbf{Ours} & \textbf{25.454} & \textbf{0.7969} & \textbf{0.0043} & \textbf{29.550} &\textbf{0.8966}& \textbf{0.0020} & \textbf{64.69}  \\
    \hline\hline
    \end{tabular}
}
\label{tab:Dataset12}
\end{table}

\begin{table}[t]
\centering
\fontsize{8}{10}\selectfont 
\caption{Quantitative comparison on Clinical Dataset.}
\resizebox{\linewidth}{!}{  
\begin{tabular}{
>{\centering\arraybackslash}p{1.8cm}|
*{2}{c}|
*{2}{c}|
*{2}{c}|
*{2}{c}|
*{2}{c}|
*{2}{c}
}
\hline\hline
\multirow{2}{*}{\textbf{Method}} & 
\multicolumn{2}{c|}{\textbf{Case1}}& 
\multicolumn{2}{c|}{\textbf{Case2}} &
\multicolumn{2}{c|}{\textbf{Case3}} &
\multicolumn{2}{c|}{\textbf{Case4}} &
\multicolumn{2}{c|}{\textbf{Case5}} &
\multicolumn{2}{c}{\textbf{Case6}}
\\
\cline{2-13}
& \multicolumn{1}{c}{PSNR $\uparrow$} & \multicolumn{1}{c|}{SSIM $\uparrow$}
& \multicolumn{1}{c}{PSNR $\uparrow$} & \multicolumn{1}{c|}{SSIM $\uparrow$}
& \multicolumn{1}{c}{PSNR $\uparrow$} & \multicolumn{1}{c|}{SSIM $\uparrow$}
& \multicolumn{1}{c}{PSNR $\uparrow$} & \multicolumn{1}{c|}{SSIM $\uparrow$}
& \multicolumn{1}{c}{PSNR $\uparrow$} & \multicolumn{1}{c|}{SSIM $\uparrow$}
& \multicolumn{1}{c}{PSNR $\uparrow$} & \multicolumn{1}{c}{SSIM $\uparrow$}
 \\
\hline
NeRF & 18.282 & 0.3388 & 21.880 & 0.5363 & 17.329 & 0.3835 & 23.119 &0.5985&21.634&0.5980&25.376&0.6660  \\
TensoRF & 17.206 &0.3593 & 22.787 & 0.6378 & 19.999 & 0.5124 & 22.681&0.5692&22.252&0.6816&26.158&0.7389  \\
Ultra-NeRF &  17.661& 0.2404 & 18.194 & 0.3994 & 19.030 & 0.3335  &19.053&0.4331&20.648&0.4971&22.450&0.7030 \\
\hline
3DGS&  17.280& 0.4270& 20.367 & 0.5510 & 18.524 & 0.4420 &21.807  &0.6673&22.110&0.6773&25.927&0.7333 \\
SuGaR&  18.170& 0.4601& 23.831 & 0.7085 & 20.902 & 0.5128 &23.541 &0.6381&20.135&0.5923&27.264&0.7721 \\
\textbf{Ours}  & \textbf{18.846}& \textbf{0.4978} & \textbf{24.644} & \textbf{0.7380} & \textbf{21.570} & \textbf{0.5493} &\textbf{24.030}&\textbf{0.6719}&\textbf{22.723}&\textbf{0.6911}&\textbf{28.181}&\textbf{0.7888} \\
\hline\hline
\end{tabular}}
\label{tab:Case}
\end{table}

\section{Experiments}

\subsection{Datasets} 

\textbf{Ultrasound in the Wild Dataset (Wild Dataset) \cite{dagli2024nerf}:} Comprising ten knee samples acquired via a handheld Butterfly iQ+ (30 FPS), this dataset features longitudinal suprapatellar sweeps processed. The inherent motion artifacts and trajectory variations serve to evaluate algorithm robustness under realistic, uncontrolled conditions.

\begin{figure}[htbp] 
    \centering
    \includegraphics[width=0.45\textwidth]{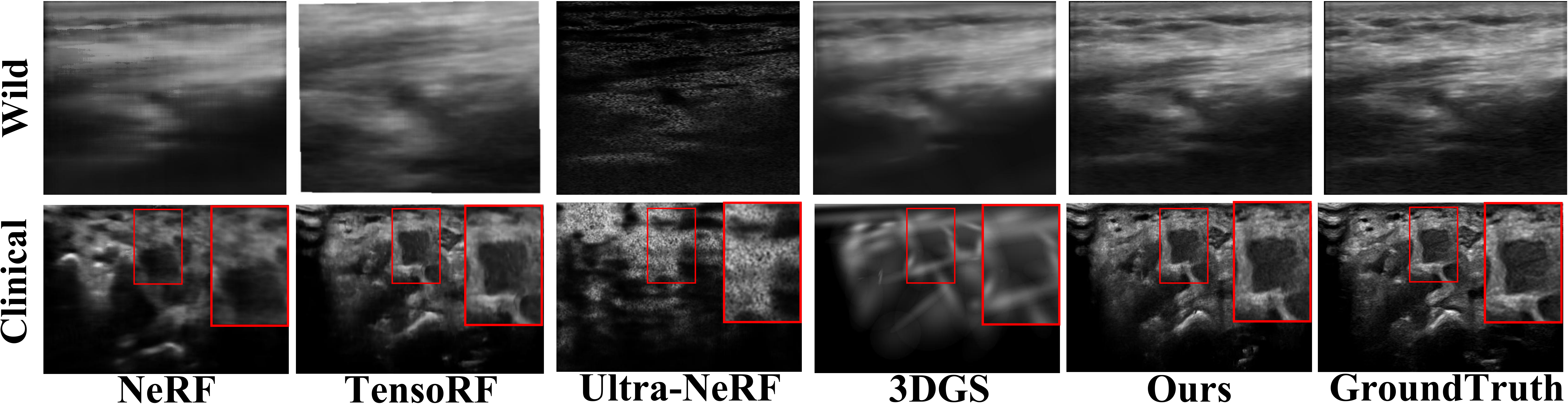}
    \caption{A visual reconstruction comparison on Wild \& Clinical Dataset.}
    \label{fig3}
    \vspace{-10pt}
\end{figure}

\textbf{Phantom Dataset \cite{wysocki2024ultra}:} This dataset consists of nine lumbar spine phantom scans acquired using a KUKA robotic manipulator. Despite precise tracking, the inclusion of tilted and perpendicular sweeps intentionally induces spinous process occlusions, enabling evaluation under controlled geometric constraints with realistic anatomical blockages.

\textbf{Clinical Ultrasound Examination Dataset (Clinical Dataset):} This is our in-house dataset, collected using the Canon i900 ultrasound system with institutional research ethics approval (Ethics Approval No.: SL2024-KY-29-01).  It consists of six challenging cases, each involving unconstrained probe motion, anatomical occlusions, and clinically realistic scanning variability. Three wrist joint cases using volar scanning for carpal tunnel visualization, and three kidney cases employing a dual-plane strategy (transverse and longitudinal) for comprehensive coverage. Acquired via standard protocols, these data reflect the inherent complexity of real-world clinical imaging. 

\begin{figure}[htpb] 
    \centering
    \includegraphics[width=0.9\columnwidth]{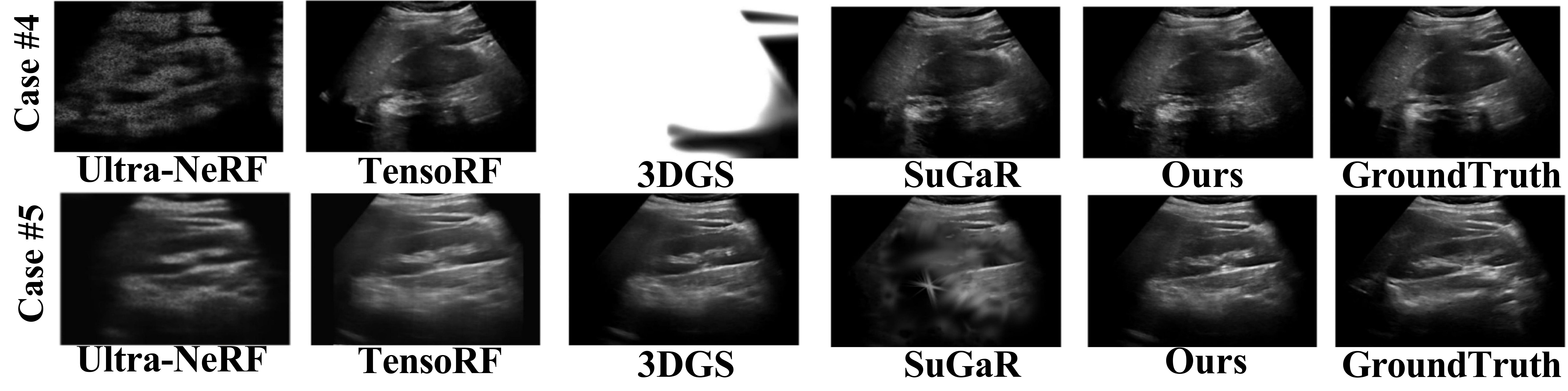}
    \caption{Comparison of kidney visualization results from Clinical Dataset.}
    \label{fig-2}
\end{figure}

\subsection{Experimental Setup} 

Following established protocols \cite{dagli2024nerf,wysocki2024ultra,guo2024ulre}, test sets were constructed by selecting every eighth frame from each dataset. We evaluated UltraGS against NeRF \cite{mildenhall2021nerf}, TensoRF \cite{chen2022tensorf}, Ultra-NeRF \cite{wysocki2024ultra}, 3DGS \cite{kerbl20233d}, and SuGaR \cite{guedon2024sugar} using average PSNR, SSIM, MSE, and inference speed. All experiments were conducted on an NVIDIA RTX 3090 GPU.

\begin{figure}[htpb] 
    \centering
    \includegraphics[width=0.4\textwidth]{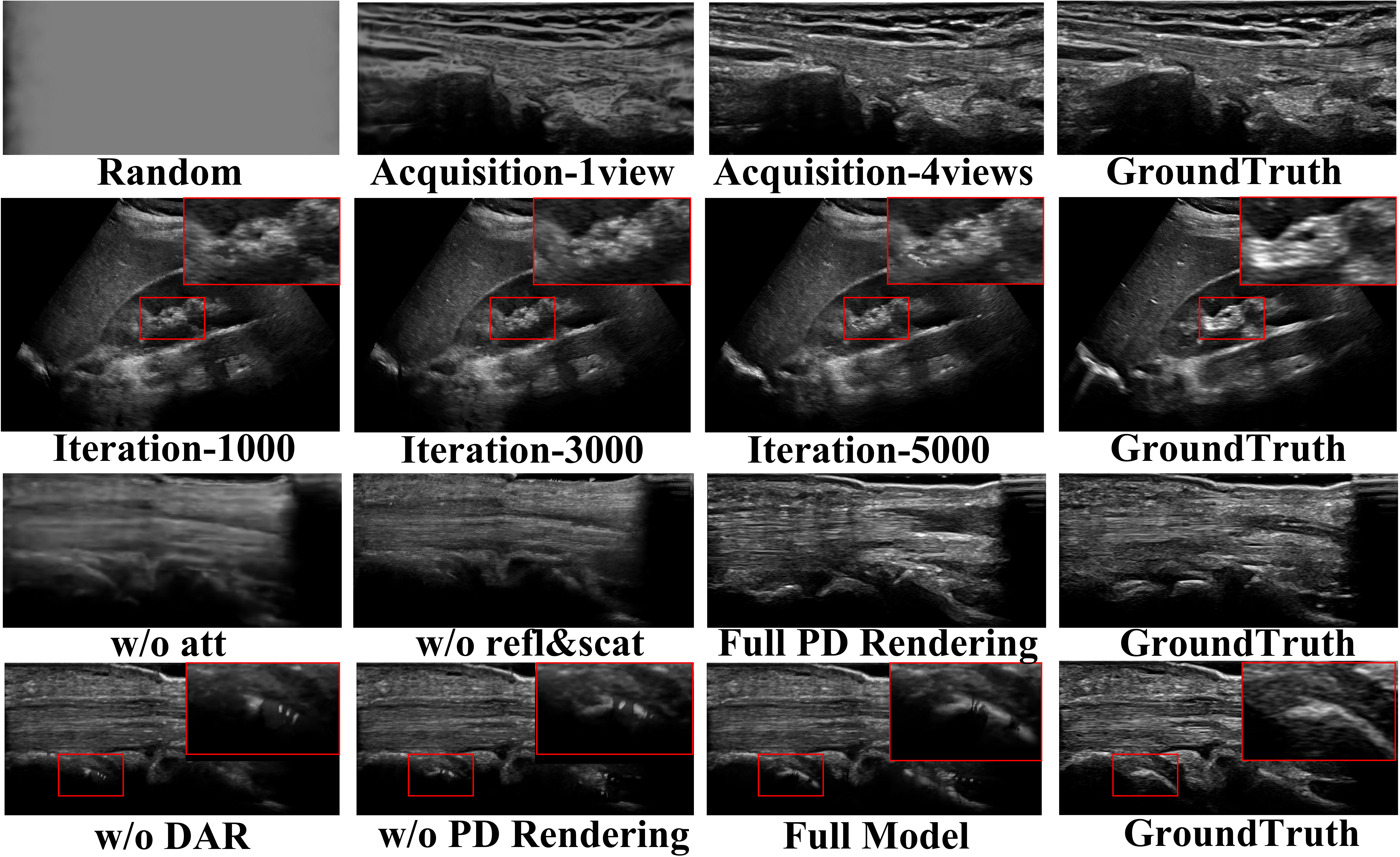}
    \caption{A visual reconstruction comparison on Ablation Study.}
    \label{fig4}
\end{figure}

\subsection{Quantitative and Qualitative Evaluation}
Table \ref{tab:Dataset12} compares UltraGS with state-of-the-art implicit and explicit radiance field methods on the Wild and Phantom datasets. UltraGS consistently achieves the best performance across all metrics, yielding the highest PSNR and SSIM and the lowest MSE. On the Wild dataset with unconstrained handheld motion, UltraGS attains a PSNR of 25.454 dB, surpassing 3DGS (22.327 dB) and NeRF-based methods. Moreover, UltraGS achieves real-time inference at 64.69 fps on an NVIDIA RTX 3090 GPU, which is an order of magnitude faster than TensoRF and Ultra-NeRF. These results demonstrate that combining explicit splatting with PD Rendering effectively eliminates the latency of volume ray marching while preserving reconstruction fidelity.

Results on the Clinical Ultrasound Examination dataset are summarized in Table \ref{tab:Case}. UltraGS outperforms all competing methods across six anatomical cases, including carpal tunnel and renal scans. NeRF and Ultra-NeRF suffer from blurred reconstructions due to strong speckle noise, while standard 3DGS exhibits structural misalignment. In contrast, UltraGS consistently improves PSNR and SSIM, achieving up to 28.181 dB PSNR in Case 6. Compared with SuGaR, which incorporates surface-alignment priors, UltraGS remains more robust across varying tissue depths, highlighting the benefit of DAR for sensorless clinical sweeps.

Qualitative comparisons in Fig. \ref{fig3} and Fig. \ref{fig-2} corroborate these findings. NeRF-based methods show pronounced volumetric blurring, and 3DGS produces center-offset artifacts in depth maps. While TensoRF and SuGaR yield sharper edges in isolated cases, their performance degrades under unconstrained conditions. UltraGS better preserves sharp anatomical boundaries and tissue–probe interfaces by explicitly modeling wave–matter interactions through the PD rendering operator. Additional qualitative results and real-time demonstrations are provided in the supplementary material.

\subsection{Ablation Study and Robustness Analysis}
We investigate model robustness and component contributions via sparse-data constraints and module removal experiments, as visualized in Fig. \ref{fig4}.

Robustness to Data Scarcity: To assess the framework’s stability under clinical constraints, we evaluated its performance across varying viewpoints and iteration counts. As illustrated in the first two rows of Fig. \ref{fig4}, single-view acquisitions produce substantial structural blur, whereas UltraGS achieves remarkable clarity with as few as four sparse views. Moreover, the model converges rapidly, although results after 1,000 iterations remain somewhat hazy, those at 3,000 iterations already capture sharp anatomical boundaries nearly indistinguishable from the fully trained 5,000-iteration output. This high data efficiency is particularly valuable in resource-limited clinical settings, where long scanning times or dense view coverage may be impractical.

Acoustic Physics Decomposition: We decoupled the PD Rendering module to evaluate the contribution of individual wave–matter interaction terms. As shown in the third row of Fig. \ref{fig4} and Table \ref{tab:ablation_limited}, omitting attenuation ($I_{att}$) results in a loss of depth-dependent contrast, failing to capture the natural energy decay observed in deeper tissues. Excluding reflection and scattering ($I_{refl}, I_{scat}$) produces images with a flat appearance, lacking the characteristic speckle texture and high-contrast boundaries at tissue–probe interfaces. By incorporating all terms, the full PD Rendering yields the most physically faithful reconstructions, effectively reconciling first-order wave physics with neural SH coefficients.

Core Module Validation: We next evaluated the two core components of our framework, DAR and the PD Rendering operator. As shown in Table \ref{tab:ablation_limited} and the final row of Fig. \ref{fig4}, removing DAR leads to pronounced geometric distortions and structural ghosting caused by uncorrected initialization drift. Replacing PD Rendering with traditional high-order SH ($l=3$, as used in standard 3DGS) introduces severe intensity inconsistencies and volumetric artifacts because conventional SH, while effective for complex optical light transport, lacks the acoustic physics priors necessary to model the non-linear wave–matter interactions in ultrasound. The superior performance of the full model demonstrates that combining neural-geometric refinement with physically grounded rendering is crucial for achieving metric-consistent 3D ultrasound reconstruction.

\begin{table}[H]
\centering
\footnotesize
\vspace{-10pt}
\caption{Ablation Study on Module Removal.}
\begin{tabular}{
  >{\centering\arraybackslash}p{3cm}|
  *{2}{c}|
  *{2}{c}
}
\hline\hline
\multirow{2}{*}{\textbf{Variant}} &
\multicolumn{2}{c|}{\textbf{Wild Dataset}} &
\multicolumn{2}{c}{\textbf{Phantom Dataset}} \\
\cline{2-5}
& PSNR & SSIM 
& PSNR & SSIM \\
\hline
w/o $I_{att}$ & 24.115 & 0.7732 & 28.230 & 0.8641 \\
w/o $I_{refl} \& I_{scat}$ & 24.380 & 0.7801 & 28.512 & 0.8715 \\
\hline
w/o DAR  & 23.422 & 0.7520 & 27.105 & 0.8500  \\
w/o PD Rendering& 23.887 & 0.7610 & 27.451 & 0.8320  \\
\hline
Ours   (Full model)     & \textbf{25.454} & \textbf{0.7969}  & \textbf{29.550} &\textbf{ 0.8966}  \\
\hline\hline
\end{tabular}
\label{tab:ablation_limited}
\end{table}

\section{Conclusion}
This study demonstrates that Gaussian splatting can be effectively adapted for ultrasound novel view synthesis by integrating depth-aware primitives and lightweight acoustic modeling. By implementing a primitive strategy with dynamic aperture rectification, we achieve metric-consistent 3D structural reconstruction of ultrasound data. Furthermore, the PD Rendering paradigm reconciles first-order wave physics with spherical harmonics to faithfully synthesize acoustic intensities while ensuring computational efficiency. Our framework achieves state-of-the-art results across wild, phantom, and clinical datasets while maintaining an unprecedented real-time performance of 64.69 fps. This highlights the efficiency and robustness of our neural-geometric joint optimization, offering a promising, hardware-free solution for clinical ultrasound applications.

\bibliographystyle{IEEEbib}
\bibliography{references}

\appendix

\section*{Supplementary Material for UltraGS}

\addcontentsline{toc}{section}{Supplementary Material for UltraGS}

\section{Mathematical Proxy}
\subsection{Justification of the Virtual Pinhole Proxy}
As discussed in the main paper, ultrasound imaging fundamentally differs from optical imaging in that it relies on three-dimensional planar slicing rather than volumetric projection. Within our sensorless reconstruction paradigm, the virtual pinhole model is therefore introduced strictly as a mathematical proxy rather than a physical representation of the ultrasound acquisition process. This abstraction is necessitated by the absence of external tracking hardware, which requires a differentiable coordinate parameterization to enable end-to-end optimization \cite{adriaans2024trackerless}. 

Specifically, we adopt the virtual pinhole formulation to provide an initial projective scaffold upon which optimization can proceed. Stochastic anchors obtained via structure-from-motion serve only as a coarse initialization. Subsequently, the proposed neural–geometric feedback loop progressively compensates for the discrepancy between projective geometry and acoustic image formation through joint optimization of Gaussian primitives and learnable field-of-view parameters. As a result, the system is able to reconcile metric inconsistencies introduced by the proxy model and converge toward an acoustically plausible spatial configuration.

\subsection{COLMAP for Ultrasound: Rationale and Pre-processing}
Although COLMAP \cite{schoenberger2016sfm} is originally developed for optical imagery, its hardware-agnostic design makes it a practical initialization tool in clinical settings where probe tracking is unavailable. Direct application to ultrasound data, however, is challenged by low contrast, speckle noise, and reduced feature repeatability. To address these limitations, we apply targeted pre-processing prior to affine SfM initialization.

In particular, anisotropic diffusion filtering is employed to suppress speckle noise while preserving structural edges, thereby stabilizing keypoint detection. In addition, contrast-limited adaptive histogram equalization is applied to enhance local contrast in low signal-to-noise regions, improving feature matching robustness across views. These steps ensure that COLMAP yields a sufficiently stable coordinate baseline, which is subsequently refined by UltraGS into a metric-consistent volumetric representation.

\subsection{Initialization Sensitivity Analysis}
Empirical results indicate that implicit reconstruction methods, such as Ultra-NeRF \cite{wysocki2024ultra}, exhibit strong sensitivity to the accuracy of pose initialization. When high-precision robotic poses are replaced by COLMAP-derived anchors, these implicit architectures frequently fail to converge or suffer from severe structural blurring. This behavior can be attributed to the limited geometric degrees of freedom available for correcting early-stage misregistrations.

In contrast, UltraGS demonstrates substantially greater robustness to initialization noise. By leveraging explicit two-dimensional Gaussian primitives together with Dynamic Aperture Rectification, the model retains sufficient flexibility to correct pose drift and focal inconsistencies during optimization. This explicit geometric representation enables the reconstruction process to gradually self-correct initial errors and preserve structural integrity throughout training.

\section{Implementation Details}

\subsection{Optimization Constraints}
To maintain physical plausibility within the proposed PD rendering paradigm, we impose explicit constraints during optimization. All learnable acoustic weights, including attenuation, reflection, and scattering coefficients, are constrained to remain positive through appropriate activation functions, ensuring compliance with acoustic energy conservation principles. 

Furthermore, the learnable field-of-view parameters are initialized according to probe-specific specifications and constrained within bounded ranges. This prevents degenerate focal configurations and stabilizes the Dynamic Aperture Rectification process during training.

\subsection{Hyperparameters}
The model is trained for a total of 5000 iterations using the Adam optimizer. The learning rate for Gaussian position updates is set to $1.6 \times 10^{-4}$, while the field-of-view parameters are optimized with a higher learning rate of $1.0 \times 10^{-3}$. This asymmetric schedule facilitates rapid geometric rectification in the early stages of training while maintaining stable convergence of spatial primitives.

\section{Visualization Results}

\begin{figure}[H] 
    \vspace{-15pt}
    \centering
    \includegraphics[width=0.45\textwidth]{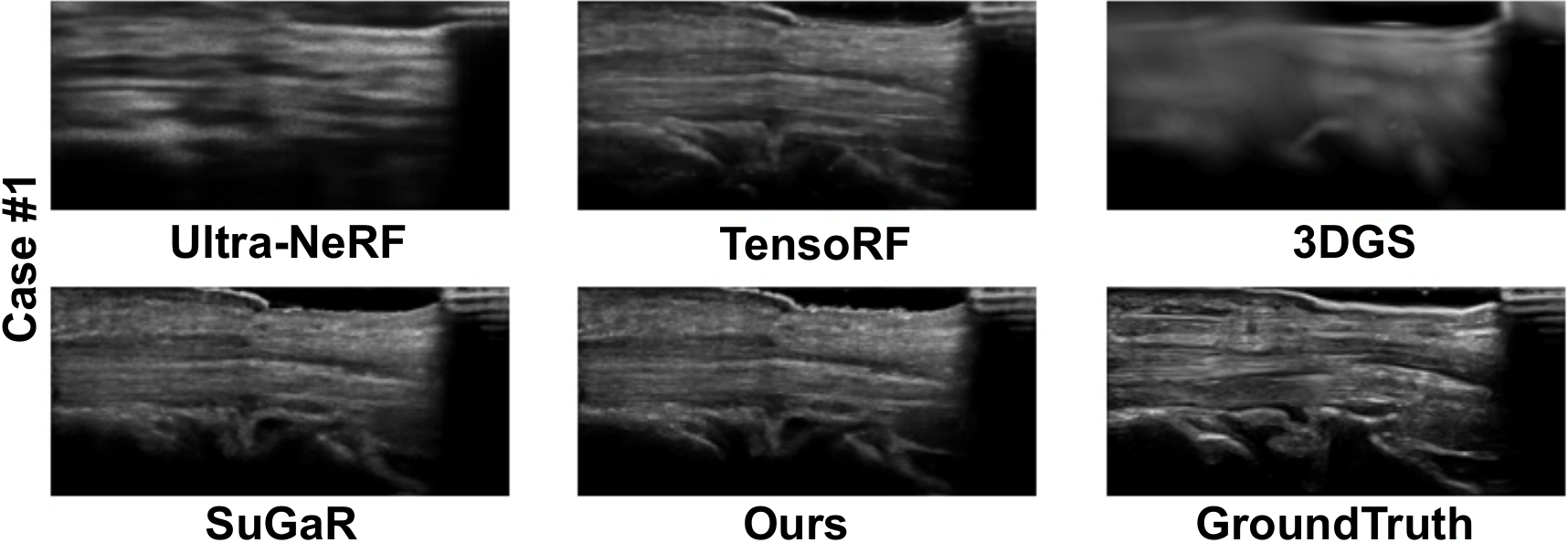}
    \caption{Visual Comparison for Case 1 in the Clinical Dataset.}
    \label{case1}
    \vspace{-20pt}
\end{figure}

\begin{figure}[H]  
    \centering
    \includegraphics[width=0.45\textwidth]{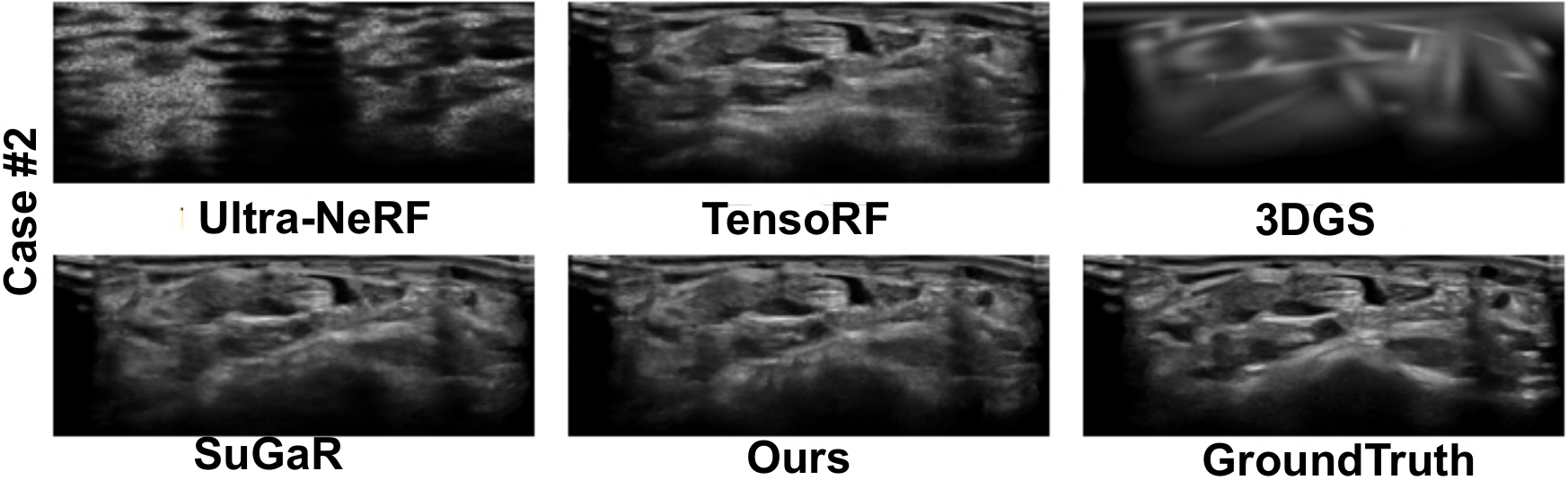}
    \caption{Visual Comparison for Case 2 in the Clinical Dataset.}
    \label{case2}
\end{figure}

\begin{figure}[H]  
    \centering
    \vspace{-10pt}
    \includegraphics[width=0.45\textwidth]{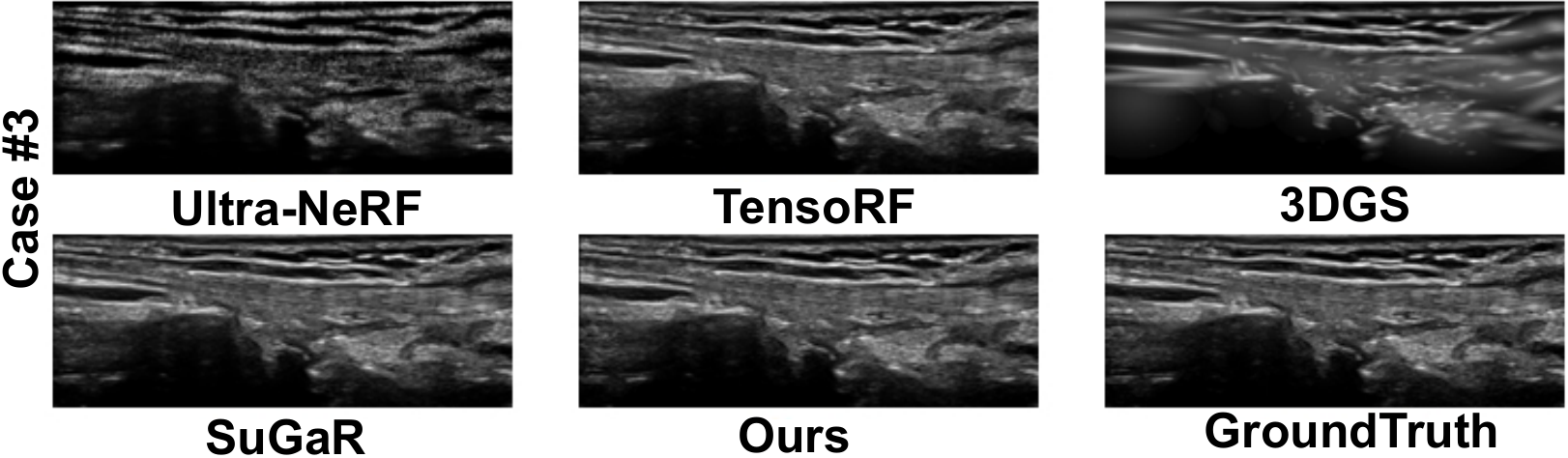}
    \caption{Visual Comparison for Case 3 in the Clinical Dataset.}
    \vspace{-20pt}
    \label{case3}
\end{figure}

\begin{figure}[H]  
    \centering
    \includegraphics[width=0.45\textwidth]{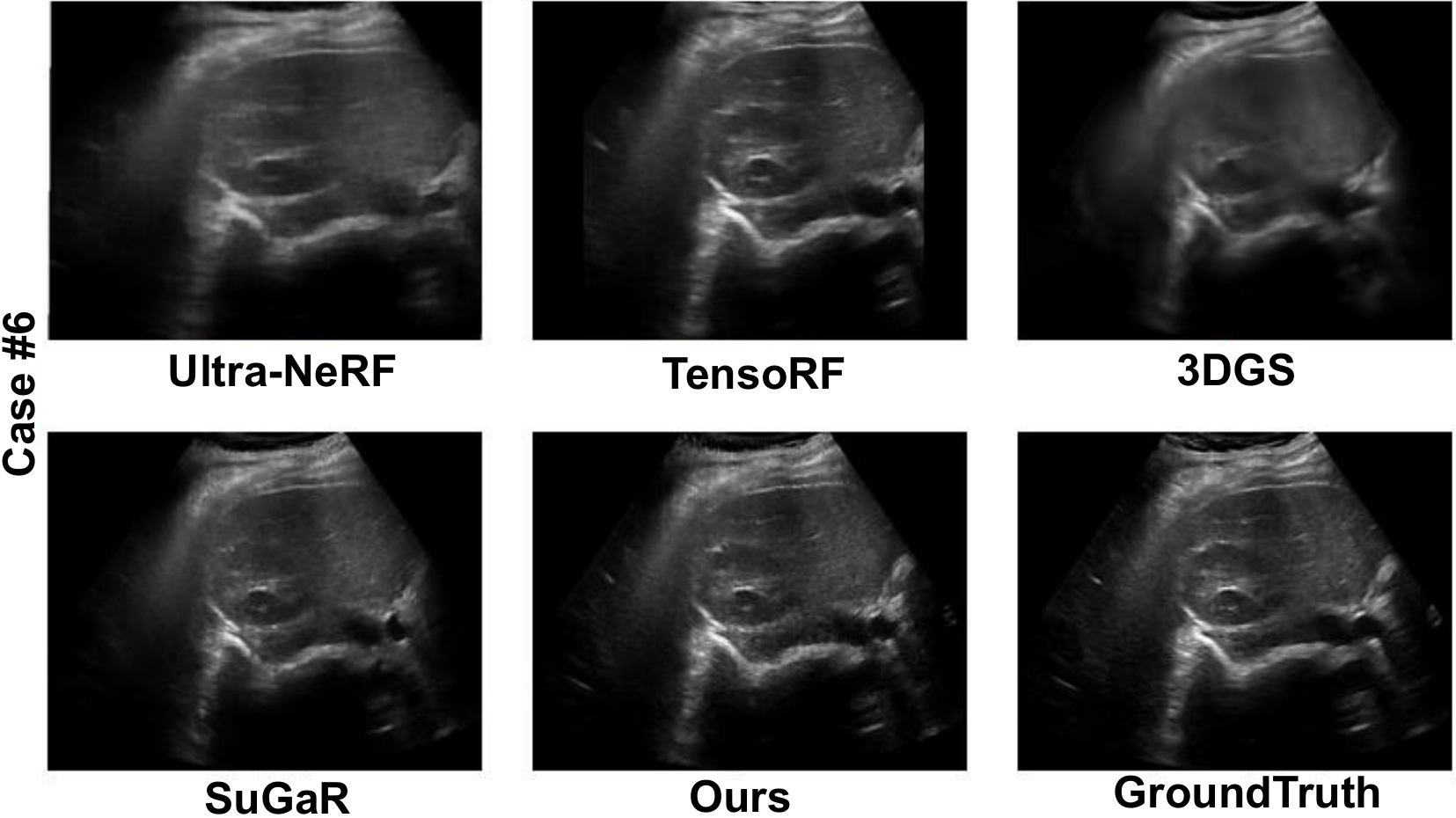}
    \caption{Visual Comparison for Case 6 in the Clinical Dataset.}
    \label{case6}
\end{figure}

\begin{figure}[H]  
    \vspace{-10pt}
    \centering
    \includegraphics[width=0.4\textwidth]{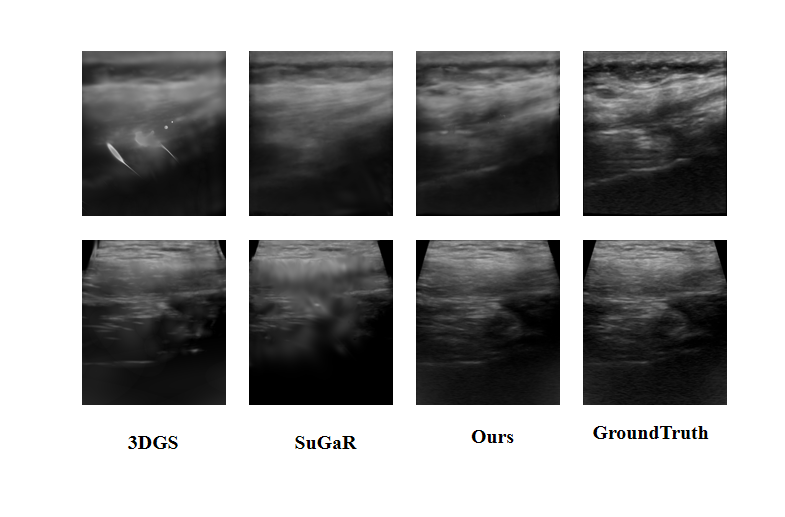}
    \caption{Visual Comparison for Wild Dataset.}
    \label{1}
    \vspace{-10pt}
\end{figure}

\begin{figure}[H]  
    \vspace{-10pt}
    \centering
    \includegraphics[width=0.4\textwidth]{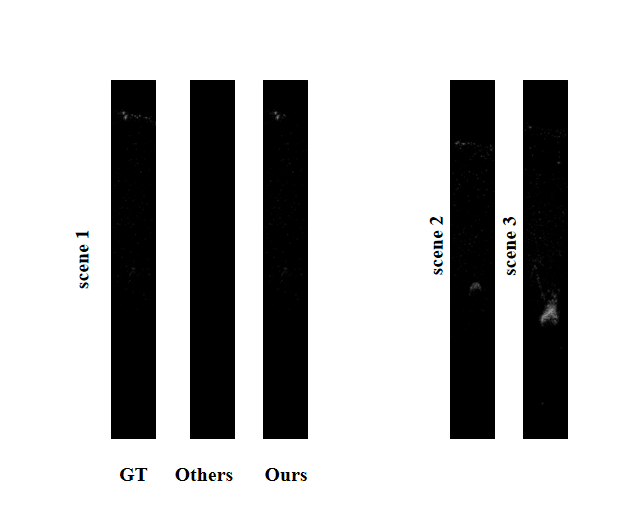}
    \vspace{-10pt}
    \caption{Visual Comparison for Phantom Dataset.}
    \label{2}
\end{figure}

As shown in Fig. \ref{case1}-\ref{case6}, UltraGS consistently preserves sharp anatomical boundaries even in complex renal and wrist scans. The visual results for the Wild and Phantom datasets (Fig. \ref{1}-\ref{2}) further demonstrate that our framework avoids the volumetric blurring common in NeRF-based baselines while maintaining higher structural contrast than standard 3DGS \cite{kerbl20233d}.

\section{Limitations and Future Work}
The current PD Rendering module utilizes first-order approximations of wave physics. While efficient, it may not fully capture high-order reverberation artifacts in certain pathological conditions.Future work will focus on integrating more complex acoustic priors and developing a fully ultrasound-specific SfM framework to further enhance registration robustness in unconstrained clinical environments.

\end{document}